\pdfoutput=1

\documentclass[11pt]{article}

\usepackage{EMNLP2022}

\usepackage{times}
\usepackage{latexsym}

\usepackage[T1]{fontenc}

\usepackage[utf8]{inputenc}

\usepackage{microtype}

\usepackage{inconsolata}

\usepackage{amsmath}
\usepackage{graphicx}
\usepackage{mathtools,amssymb}
\usepackage{multirow, makecell}
\usepackage{multicol}
\usepackage{booktabs}
\usepackage{subcaption}
\usepackage{CJKutf8}
\usepackage{color}
\usepackage{paralist}
\usepackage{xspace}

\newcommand{\canmt}{CANMT\xspace}
\newcommand{\enc}{encoder\xspace}
\newcommand{\dec}{decoder\xspace}
\newcommand{\est}{self-estimator\xspace}
\title{Competency-Aware Neural Machine Translation:\\ Can Machine Translation Know its Own Translation Quality?}

\author{Pei Zhang~\thanks{*Corresponding author.}, Baosong Yang, Haoran Wei, Dayiheng Liu, Kai Fan, Luo Si \and Jun Xie \\
Alibaba Group Inc.\\
\texttt{\{xiaoyi.zp,yangbaosong.ybs,funan.whr,qingjing.xj\}@alibaba-inc.com}}
\begin{document}
\maketitle
\begin{abstract}

Neural machine translation (NMT) is often criticized for failures that happen
without awareness. 
The lack of competency awareness makes NMT untrustworthy. 
This is in sharp contrast to human translators who give feedback or conduct further investigations whenever they are in doubt about predictions. 
To fill this gap, we propose a novel competency-aware NMT by extending conventional NMT with a self-estimator, offering abilities to translate a source sentence and estimate its competency.
The self-estimator encodes the information of the decoding procedure and then examines whether it can reconstruct the original semantics of the source sentence. 
Experimental results on four translation tasks demonstrate that the proposed method not only carries out translation tasks intact but also delivers outstanding performance on quality estimation.
Without depending on any reference or annotated data typically required by state-of-the-art metric and quality estimation methods, our model yields an even higher correlation with human quality judgments than a variety of aforementioned methods, such as BLEURT, COMET, and BERTScore. 
Quantitative and qualitative 
analyses show better robustness of competency awareness in our model.\footnote{Code and test sets are available at: https://github.com/xiaoyi0814/CANMT.}

\end{abstract}

\section{Introduction}

With the exploitation of large amounts of training data, neural machine translation (NMT) models~\cite{Bahdanau2015NeuralMT,vaswani2017attention} have shown promising quality. 
However, NMT often makes mistakes, especially on outliers such as out-of-domain samples, input noises, and low-frequency words, on which models are potentially not trained or evaluated~\cite{Koehn2017SixCF, Khayrallah2018OnTI, Mller2020DomainRI}.  
Unlike human translators who are generally aware of their expertise and weaknesses, NMT lacks competency awareness, which limits the trustworthiness when it is widely adopted.
Consequently, it is necessary to equip NMT models with good awareness of their translation competency, i.e. the ability to give a translation along with the quality score of the translation. 
With good awareness of its competency, NMT can then take conservative actions (such as human intervention) on low-quality translations, and request fewer manual inspections for high-quality ones.

\begin{table}
\centering
\small
\begin{tabular}{lcc}
\hline
\bf {Method} &
\bf{Input} $\to$ \bf{Output} &
\bf{Human-Assist} \\
\hline
MT & \textit{(src) $\to$ (trans)} & N/A \\
Metric   &  \textit{(src, ref, trans) $\to$ (quality)} & \checkmark \\
QE       &  \textit{(src, trans) $\to$ (quality)} &\checkmark\\
CANMT & \textit{(src) $\to$ (trans, quality)} & \texttimes \\
\hline
\end{tabular}

\caption{Comparison among Machine Translation (MT), Automatic Translation Evaluation (Metric), Quality Estimation (QE), and our Competency-Aware Neural Machine Translation (CANMT). `Input$\to$Output' indicates the input and output for each task, where \textit{src, ref, trans}, and \textit{quality} represent source, reference, translation, and quality score respectively. `Human-Assist' denotes whether quality evaluation depends on human-annotated references or samples with quality scores for training.}
\label{tab:differ}
\end{table}

In order to inform users of the competency of MT, a traditional alternative is to employ extra quality evaluation tools\footnote{For simplification, we call these methods as extra-estimation tools in the subsequent of this paper.} that are independent of the MT model, including automatic evaluation metrics~\citep[Metric;][]{papineni2002bleu,rei2020comet,Wan2021RoBLEURTSF} and quality estimation~\citep[QE;][]{Kim2016ARN, Fan2019BilingualEC}.
As shown in Table~\ref{tab:differ}, given a source sentence and machine translation, metrics evaluate quality score relying on reference translations, thus being unsuitable for tackling this issue. 
Unlike metrics that require references, QE estimates the quality of translation given only the source sentence~\cite{Kim2016ARN}. 
Nevertheless, well-performing QE models are typically supervised methods trained on a large amount of expert-annotated data~\cite{specia-etal-2020-findings-wmt, Specia2021FindingsOT}.
Besides, recent studies have pointed out that the extra-estimation tools like metrics and QE potentially drop accuracy when switching to out-of-domain~\cite{Freitag2021ResultsOT}.

In this paper, we propose a novel method, \textbf{C}ompetency-\textbf{A}ware \textbf{N}eural \textbf{M}achine \textbf{T}ranslation (\textbf{CANMT}), which has both translation and self-estimation capabilities without any ground-truth translation or expert-scored data for the competency assessment.
Based on the assumption that ``a good translation should be able to reconstruct the meaning of source sentence from it'', we 
extend the standard translation model with a self-estimator, which adopts a reconstruction strategy, and use the semantic gap between the original source sentence and its reconstruction to achieve competency awareness.
Our self-estimator is a part of the NMT model and exploits the continuous representations as input.
The intuition we believe is that continuous representations are more informative about the decoding procedure of NMT, thus better reflecting the quality of translation candidates than the discrete output. 
Moreover, we update the original NMT decoder to a two-stream decoder, in which translation and reconstruction information flows are separated to avoid the interference of the source sentence during reconstruction.

We conduct experiments on four widely used translation tasks: CWMT17 Chinese-to-English (Zh$\to$En), ASPEC Japanese-to-English (Ja$\to$En), WMT14 French-to-English (Fr$\to$En) and WMT14 English-to-German (En$\to$De).
Results show that while keeping the same translation performance as Transformer~\cite{vaswani2017attention} baseline, CANMT exhibits promising capability on self-estimation of quality, which outperforms the state-of-the-art unsupervised metrics and QE methods such as BERTScore~\cite{zhang2019bertscore}, RTT~\cite{Moon2020RevisitingRT} and confidence-based QE~\cite{fomicheva2020unsupervised}. It is encouraging to see that our method even surpasses those existing supervised extra-estimation tools like BLEURT~\cite{sellam2020bleurt} and COMET~\cite{rei2020comet} which depend on ground-truth translations or learn from expert-scored data. 
Extensive analyses demonstrate the robustness of CANMT that effectively tackles problems of out-of-domain, quality drift, and miscalibration in self-estimation. 
Complementary effects are also observed between CANMT and extra-estimation tools.

\begin{figure*}[t!]
\centering
\includegraphics[width=0.85\textwidth]{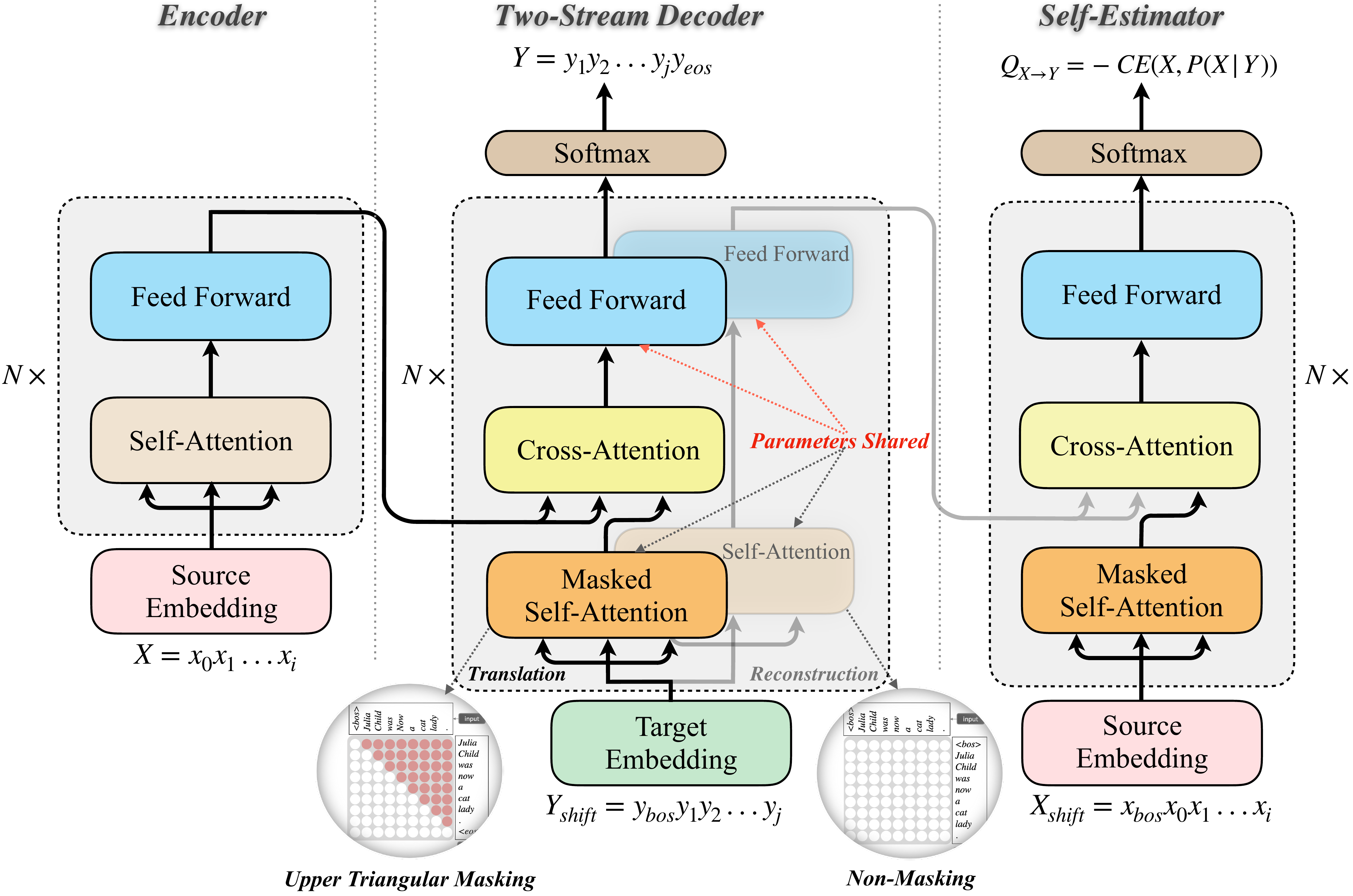}
\caption{
Illustration of CANMT, where the encoder-decoder NMT model is extended with a self-estimator based on reconstruction. We introduce a two-stream mechanism in the decoder for both translation and reconstruction. A source sentence $X$ is first encoded, and then joint-attended by \textit{Translation} stream of the decoder to generate translation $Y$. The continuous representation from \textit{Reconstruction} stream of the decoder is further fed into the self-estimator to reconstruct $X$. Then 
competency is estimated by the cross-entropy (CE) loss of reconstruction. The difference in self-attention masks for the two-stream decoder is elaborated in the circular zoom areas.
}
\label{fig:model}
\end{figure*}

\section{Competency-Aware Neural Machine Translation (CANMT)}

Given an arbitrary source sentence $X$, the goal of \canmt model is to generate a target translation $Y$ and assign a self-estimated competency score $Q_{X\to Y}$ for the generated translation $Y$ without any reference translation or expert-annotated data, which can be represented as:
\begin{align}
\label{object}
Y, Q_{X\to Y} = \text{CANMT}(X).
\end{align}

We propose to exploit the reconstruction strategy and use the semantic gap between original source sentence and its reconstruction to enable competency awareness. 
The main challenge of this approach is how to capture as many quality issues of the predicted translation as possible in reconstructing, such as semantic or syntactic errors and disfluency.
To overcome this challenge, we utilize the continuous representation from the NMT decoder, as it contains more information about the decoding procedure than a discrete translation and thus is easier to reconstruct the source sentence.

\subsection{Model Architecture}

The architecture of \canmt is illustrated in Figure~\ref{fig:model}. Our model consists of three components: \textbf{\enc}, \textbf{two-stream decoder}, and \textbf{\est}, which are all built with Transformer blocks~\cite{vaswani2017attention}. 
The source sentence $X$ is first encoded by the encoder, then passed to the two-stream decoder. One stream jointly attends to source representations to generate a target translation $Y$. The other feeds the information about the decoding procedure to the self-estimator, which then reconstructs the source sentence and estimates quality using cross-entropy loss. 
In the following, we specify the structure of each component.

\paragraph{Encoder}
The encoder of \canmt is the same as the standard Transformer. It maps the token sequence of input sentence $X = (x_1,...,x_l)$ (with length of $l$) to a contextual representation $C_X$:
\begin{align}
\small
    C_X = \text{Encoder}([e(x_1),...,e(x_l)]; \theta_{Enc}),
\end{align}
where $e(\cdot)$ is the source embedding lookup table and $\theta_{Enc}$ denotes the encoder parameters. Then $C_X$ is fed into the \dec to generate translation.

\paragraph{Two-Stream Decoder}

Since we propose to use the information of translation procedure for source-side reconstruction, however, the cross-attention in regular decoder introduces source-side information which makes the learning of reconstruction ineffective.
Therefore, we adopt a controllable two-stream mechanism in the decoder.
Given a target translation $Y = (y_1,...,y_m)$ (with length of $m$),
\begin{itemize}
    \item \textit{Translation} stream is in charge of forward translation, which serves a similar role in the vanilla Transformer. This stream can have access to the sequential information $y_{<t}$ of each target word $y_t$ and the source context. 
    \item \textit{Reconstruction} stream is in charge of the source-side reconstruction. This stream instead captures the full contextual information $y_{1:m}$ but does not include the source context to avoid information leaking.
\end{itemize}
We share the parameters in the two-stream mechanism to retain more information about the translation decoding procedure for reconstruction.

Computationally, we denote the representation of $t$-th layer in \textit{Translation} and \textit{Reconstruction} streams as $h$ and $g$, respectively. The first layer representations of the two streams are both set as the precedent word embedding, i.e. $h^{0}_t = g^{0}_t = e(y_t)$. For each layer $n = 1, 2, ..., N$, the representations of two streams are calculated as follows: \footnote{Here, we omit the computation of LayerNorm and residual connections for simplification.}
\begin{itemize}
\addtolength\itemsep{-4mm}
\item for \textit{Translation} stream
\begin{align}
\small
\begin{split}
   & \hat{h}^{(n)}_{t} =  \text{MHA}^{(n)}_{\text{slf}}(Q{=}h^{(n-1)}_t,KV{=}h^{(n-1)}_{\textcolor{blue}{<t}}) \\
   &  h^{(n)}_{t} =  \text{FFN}^{(n)}(\text{MHA}^{(n)}_{\text{ctx}}(Q{=}\hat{h}^{(n)}_{t}, KV{=}C_X))
\end{split}
\end{align}
\item for \textit{Reconstruction} stream
\begin{align}
\small
\begin{split}
&\hat{g}^{(n)}_{t} =\text{MHA}^{(n)}(
Q{=}g^{(n-1)}_t,KV{=}g^{(n-1)}_{\textcolor{blue}{1:m}}) \\
& g^{(n)}_{t} = \text{FFN}^{(n)}(\hat{g}^{(n)}_{t})
\end{split}
\end{align}
\end{itemize}
where $\text{MHA}^{(n)}_{\text{slf}}$,  $\text{MHA}^{(n)}_{\text{ctx}}$ and $\text{FFN}^{(n)}$ are self-attention, cross-attention and position-wise feed-forward sub-layers in the $n$-th decoder layer. As the causal attention masking (normally upper triangular masking) in the self-attention of the Transformer decoder would limit the contextual modeling ability and undermine the reconstruction performance, we discard this mask in \textit{Reconstruction} stream to enable the full access of context. A specific example illustrating the difference in masking mechanism for two streams is presented in the circular zoom areas in Figure~\ref{fig:model}.

The output of \textit{Translation} stream is used to generate the final translation. We use a learned linear transformation and the softmax function to convert $h^{(N)}_{i}$ to the probability $P(y_i|y_{<i}, X)$ of the $i$-th token, and obtain translation $Y$ by maximizing the product of these probabilities:
\begin{align}
\small
    Y = \text{argmax} \prod_{t=1}^{m}P(y_i|y_{<i},X).
\end{align}
Specifically, the translation inference is performed by the commonly used beam search.
The output of \textit{Reconstruction} stream, which is denoted as $C_{Y} = [g^{(N)}_{1},...,g^{(N)}_{m}]$, is fed into the \est to obtain the self-estimation score of $Y$.

\paragraph{Self-Estimator}

The structure of the self-estimator is the same as that of the regular decoder. 
In the inference time, the hidden states of \textit{Reconstruction} stream of decoder $C_{Y}$ are jointly attended by self-estimator to reconstruct the source sentence, and the output probability $P(X|Y) = \prod_{i=1}^{l} P(x_i|x_{<i}, Y)$ is computed by forced decoding with $X$.

The quality score of translation can be deduced by the negative Cross-Entropy (CE) loss between the output probability of \est and $X$, which can be mathematically represented as: 
\begin{align}
\small
Q_{X\to Y} =& - \text{CE}(X, P(X|Y)) \nonumber\\
=& \frac{1}{l}\sum_{i=1}^l \log P(x_i|x_{<i}, Y).
\end{align}
For any given example, the higher the cross-entropy is, the poorer the quality of the corresponding translation, which reflects the inability in this case.

\subsection{Training}
The training of CANMT relies entirely on a bilingual corpus and aims at joint learning of forward translation and backward reconstruction.
Let $\mathcal{D} = \{(X^n, Y^n)\}^{|\mathcal{D}|}_{n=1}$ denote the training corpus with $\mathcal{|D|}$ paired sentences, where $(X^n, Y^n)$ is a pair of sentences in source and target languages.
The final training objective is to minimize the following cross-entropy loss function:
\begin{align}
    \mathcal{L} =& \mathbb{E}_{(X^n,Y^n)\sim\mathcal{D}}[-\log P(Y^n|X^n;\theta_{Enc},\theta_{Dec}) \nonumber\\
    - & \log P(X^n|Y^n;\theta_{Dec},\theta_{Est})],
\end{align}
where $\theta_{Enc},\theta_{Dec},\theta_{Est}$ indicate the parameters of encoder, decoder and self-estimator. The two negative log-likelihoods respectively represent the training objectives of forward translation and backward reconstruction.
Significantly, the input to decoder during training is ground truth, while the input of that during inference is the model translation.  

\section{Experiments}
For the competency awareness of CANMT, the evaluation of performance consists of two aspects: translation and self-estimation.
We use BLEU to validate translation performance.
As for self-estimation, it considers whether an MT model is aware of its own competency, thus public metric and QE tests are unsuitable for us as the translations are from other models.
Therefore, we generate translations by CANMT and evaluate the Pearson correlations between the self-estimation scores and human judgments.

\begin{table*}[h]
\centering
\small
\begin{tabular}{llccccccc}
\toprule
\bf{Tasks} & \bf{Methods} & \bf{Ref.} &
\bf Zh$\to$En &
\bf Fr$\to$En &
\bf Ja$\to$En &
\bf En$\to$De &
\bf Average \\
\hline
\multirow{2}{*}{\bf{MT}} & Transformer  &  & 23.1 &  36.1 &   28.9 &  26.9 & 28.8 \\
& CANMT (Ours)  &  & 23.5 &  35.8 &  29.0 & 26.8 & 28.8 \\
\midrule
\multirow{11}{*}{\bf{Eval.}}& \multicolumn{6}{c}{Supervised Methods} \\\cline{2-8}
& COMET-QE~\cite{rei2020comet}  & \texttimes & 0.51 & \bf 0.54 &  0.13 &  0.57 & 0.44\\
& BLEURT~\cite{sellam2020bleurt}   & \checkmark & 0.46 & 0.49&	0.28&	0.31 & 0.39 \\
& COMET~\cite{rei2020comet}  & \checkmark & 0.49 & 0.50&  0.35 &	\bf 0.58 & 0.48 \\\cmidrule{2-8}
& \multicolumn{6}{c}{Unsupervised Methods} \\\cline{2-8}
& RTT-SentBLEU~\cite{Moon2020RevisitingRT}   &  \texttimes  & 0.22  & 0.13 & 0.20 & 0.28 & 0.21\\
& RTT-BERTScore~\cite{Moon2020RevisitingRT}   &  \texttimes & 0.21  & 0.24 & 0.30 & 0.49 & 0.31\\
& TP &  \texttimes & 0.35 & 0.41 &  0.41 & 0.41 & 0.40 \\
& D-TP($K=30$)~\cite{fomicheva2020unsupervised}  &  \texttimes  & 0.55  & 0.46 & 0.44 & 0.50 & 0.49 \\
& SentBLEU~\cite{papineni2002bleu}  & \checkmark & 0.08  &  0.15 & 0.09 & 0.23 & 0.14 \\
& BERTScore~\cite{zhang2019bertscore}  & \checkmark  & 0.37 &  0.42& 0.31	&	0.41 &0.38 \\
& CANMT (Ours) & \texttimes & \underline{\bf0.61} & \underline{0.50} &  \underline{\bf0.52}&  \underline{0.52} & \underline{\bf0.54}\\
\bottomrule
\end{tabular}
\caption{Results on four translation tasks. The performance of translation (MT) is measured by BLEU, while the performance of quality evaluation (Eval.) is assessed by Pearson correlations with human judgments.
``Ref.'' indicates whether  reference translations are needed for Eval. 
``Average'' denotes the average performance of four tasks. 
The best results of all the methods are marked in bold, and that of all the unsupervised methods are underlined. CANMT offers better quality evaluation ability on its translations than its unsupervised and supervised counterparts, without drops in translation quality.} 
\label{tab:pearson_metrics}
\end{table*}

\subsection{Experimental Setting}
\paragraph{Dataset}
We train and evaluate models on four standard machine translation datasets: CWMT17 Zh$\to$En, WMT14 Fr$\to$En, WMT14 En$\to$De of news domain and ASPEC~\cite{nakazawa2016aspec} Ja$\to$En of the scientific domain.
We segment Chinese and Japanese sentences with Jieba\footnote{https://github.com/fxsjy/jieba} and KyTea\footnote{http://www.phontron.com/kytea/} respectively, while the segmentation for the other languages is processed with Moses\footnote{https://github.com/alvations/sacremoses} tokenizer. 
The byte-pair encoding~\cite{Sennrich2016NeuralMT} for Zh$\to$En, Fr$\to$En, En$\to$De, and Ja$\to$En are trained respectively with 32k/40k/32k/16k merge operations and  source-target vocabularies are shared except for Zh$\to$En.
We follow the standard setup of valid and test sets for each task.
Statistics and details are presented in Appendix~\ref{appendix:dataset}.

\paragraph{Implementation Details}
The proposed CANMT can be exploited in both RNN and Transformer models.
We implement it based on the Transformer-based model with Fairseq~\cite{ott2019fairseq}.
Our encoder and decoder follow the hyper-parameter setting of Base Transformer~\cite{vaswani2017attention}, as well as the self-estimator, which has the same setting as the decoder.
All the experiments are applied with the share-embedding setting except for Zh$\to$En and trained on 8 NVIDIA Tesla V100 GPUs with a batch size of 4096 tokens.
We use beam search with a beam size of 8 and length penalty of 1.0 for Zh$\to$En, while for the other translation tasks we set beam size to 4 and length penalty to 0.6.
We perform early stopping on valid sets, average the last 5 checkpoints and report case-sensitive SacreBLEU~\cite{Post2018ACF} on test sets.

\paragraph{Baselines}
For translation performance, we compare CANMT with Transformer, while for self-estimation 
we compare with widely used metric and QE methods:
\begin{compactitem}
\item \textbf{SentBLEU} denotes the sentence-level BLEU that computed by SacreBLEU~\cite{Post2018ACF}.
\item \textbf{BERTScore}~\cite{zhang2019bertscore} calculates the cosine similarity between the sentence embeddings of hypothesis and reference translations based on BERT~\cite{Devlin2019BERTPO}. 
\item \textbf{BLEURT}~\cite{sellam2020bleurt} and \textbf{COMET}~\cite{rei2020comet} are two supervised regression-based metrics which are built based on pre-trained language models and then fine-tuned on human-scored data.
\item \textbf{COMET-QE}~\cite{rei2020comet} is the reference-less version of COMET.
\item \textbf{RTT-SentBLEU} and \textbf{RTT-BERTScore} \cite{Moon2020RevisitingRT} respectively report the sentence-level BLEU and BERTScore between a source sentence and a reconstruction outputted by an extra backward MT as introduced in Section~\ref{section:qe}.\footnote{
We used Google Translate as the backward translation system because it outperforms models trained on the WMT News training corpus in previous work.}

\begin{table}[h]
\centering
\resizebox{0.5\textwidth}{!}{
\begin{tabular}{lcccccc}
\toprule
\bf {Methods} &
\bf News &
\bf Subtitles &
\bf Laws &
\bf Ted  &
\bf Medical \\
\hline
\multicolumn{6}{c}{Supervised Methods} \\
\hline
COMET-QE & 0.51 & 0.41 & 0.36 & 0.39 & 0.51\\
BLEURT  & 0.46 & 0.28 & 0.60 & 0.40 & 0.45\\
COMET  & 0.49 & 0.35 & 0.57 & 0.47 & 0.48\\
\hline
\multicolumn{6}{c}{Unsupervised Methods} \\
\hline
TP  & 0.35 & 0.45 & 0.24 & 0.48 & 0.28\\
D-TP  & 0.55  & \bf0.58& 0.41& 0.54 & 0.48 \\
SentBLEU & 0.08  & 0.18 & 0.41 & 0.22 & 0.09\\
BERTScore  & 0.37  & 0.43 & 0.46 & 0.44 & 0.37 \\
CANMT(Ours)   & \bf0.61 & \bf0.58 & \bf0.66& \bf0.58 & \bf0.65\\
\bottomrule
\end{tabular}}
\caption{Pearson correlation on out-of-domain sets of Zh$\to$En translation direction. The best results of all the methods are marked in bold. CANMT performs better robustness on four unlearned domains.}
\label{tab:domain_drift}
\end{table}
\item \textbf{Translation Probability (TP)}~\cite{fomicheva2020unsupervised}: the sentence-level translation probability normalized by length.
\item \textbf{Monte Carlo Dropout Translation Probability (D-TP)}~\cite{fomicheva2020unsupervised}: the expectation of TP scores obtained by running $K{=}30$ stochastic forward passes through MT model perturbed by Monte Carlo Dropout.
\end{compactitem}

\paragraph{Human Evaluation}
\label{section:evaluation} 
We randomly select 200 sentences for each test set and translate them into the target language using CANMT models. 
Each sentence is scored by three professional translators presented with pairs of a source sentence and a translated hypothesis, following the scoring criteria ranging from 1 to 5\footnote{Please refer to Appendix~\ref{appendix:human_eval} for the scoring standard.} (larger is better) in previous work~\cite{Weng2020TowardsEF}. We use the average value given by three experts as the final human judgment.

\subsection{Main Results} \label{main_results}
The system-level BLEU results shown in the upper part of Table~\ref{tab:pearson_metrics} verify that our method holds the same translation performance as Transformer, which is in line with our expectations.
Therefore, we focus on the evaluation of self-estimation in the following.
The lower part of Table~\ref{tab:pearson_metrics} shows the Pearson correlations between the predicted scores and human judgments.
In general, CANMT performs pretty well and stably across tasks.

\paragraph{Comparison to the supervised methods.} Overall, our approach surpasses state-of-the-art methods like COMET, COMET-QE, and BLEURT.
Concretely, CANMT has advantages of 0.1 and 0.17 Pearson scores over the best supervised method on Zh$\to$En and Ja$\to$En and yields considerable results on Fr$\to$En and En$\to$De.
In the following, we will further verify the robustness of out-of-domain.
\paragraph{Comparison to the unsupervised methods.} 
CANMT outperforms all unsupervised methods by a large margin.
Reference-based metrics like BLEU and BERTScore rely on the given reference, but there are many possible correct translations for a given source.
They are good at distinguishing ``good'' translations that have a high degree of overlap with the given reference, but poor at judging translations that differ largely~\cite{Freitag2020BLEUMB}.
The improvements over RTT justify our motivation that the continuous representations from the decoder capture more information about the decoding than discrete translation, thus facilitating to quality estimation.
On the other hand, 
computing cross-entropy loss with forced decoding way further avoids the error accumulation of beam search in RTT during the reconstruction procedure.
Confidence-based QE method, i.e. D-TP, also gets good results.  Nevertheless, confidence-based methods mainly utilize the prediction distribution of NMT regardless of the fidelity of the source sentence and the latent states in the model, resulting in miscalibration between accuracy and confidence~\cite{guo2017calibration,wang2020inference}.
More analyses are conducted in Section~\ref{section:miscalibration}.

\paragraph{Robustness under Out-of-Domain}
To measure the reliability of CANMT under out-of-domain conditions, we further validate our method on four out-of-domain Zh$\to$En test sets.
Neither our model nor the other methods are trained or fine-tuned on data of these domains.
We collect the test sets from UM-Corpus, IWSLT, and Medline abstracts of the WMT21 biomedical task, and randomly sample around 200 sentences for each domain test set.
Results in Table~\ref{tab:domain_drift} show that our method can outperform other baselines, and achieve stable performance with smaller variances 
across all domain tests.
As proven by~\newcite{Freitag2021ResultsOT}, extra-estimation tools like COMET and BLEURT are proficient at domains on which they have trained or fine-tuned, but 
probably deficient at unseen domains.
On the contrary, our method performs better across out-of-domain tests due to its competency awareness. Even if the domain has not been learned by CANMT, it still knows its incapability.

\begin{figure*}
     \centering
     \begin{subfigure}[b]{0.493\textwidth}
         \centering
         \includegraphics[width=\textwidth]{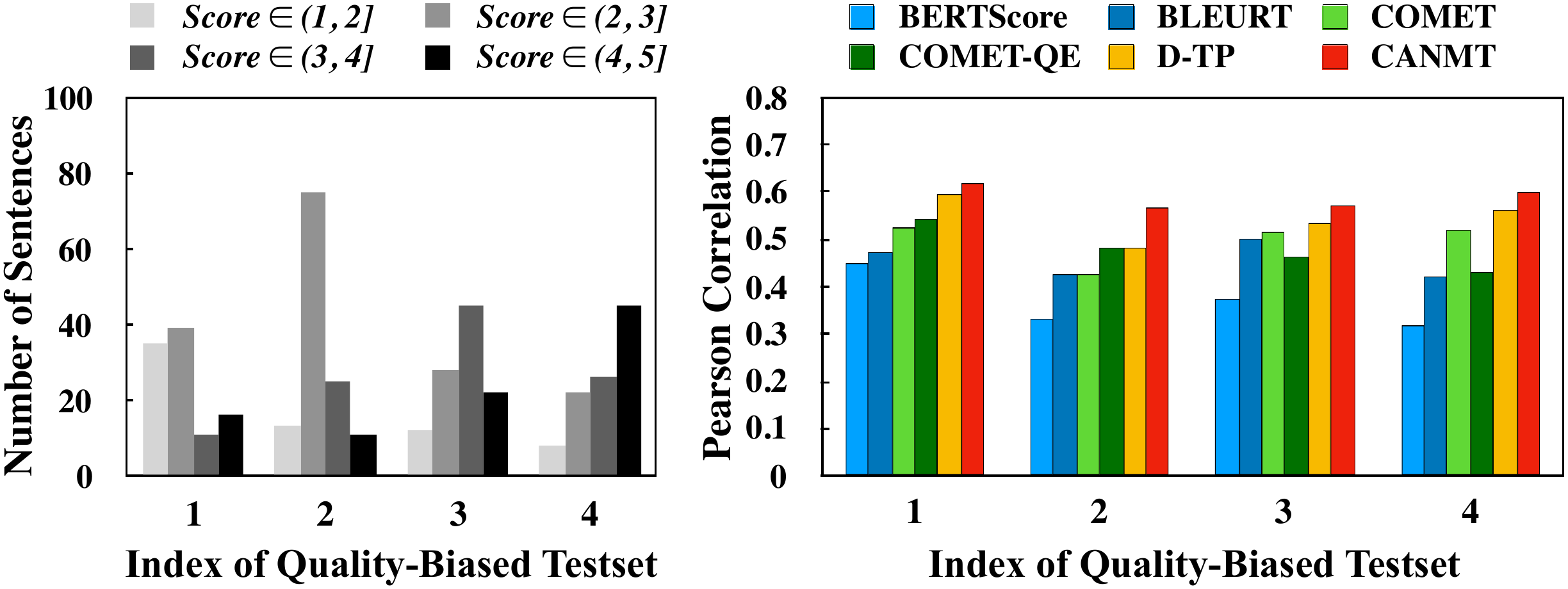}
         \caption{Results on News Domain.}
         \label{fig:quality_drift_news}
     \end{subfigure}
     \begin{subfigure}[b]{0.497\textwidth}
         \centering
         \includegraphics[width=\textwidth]{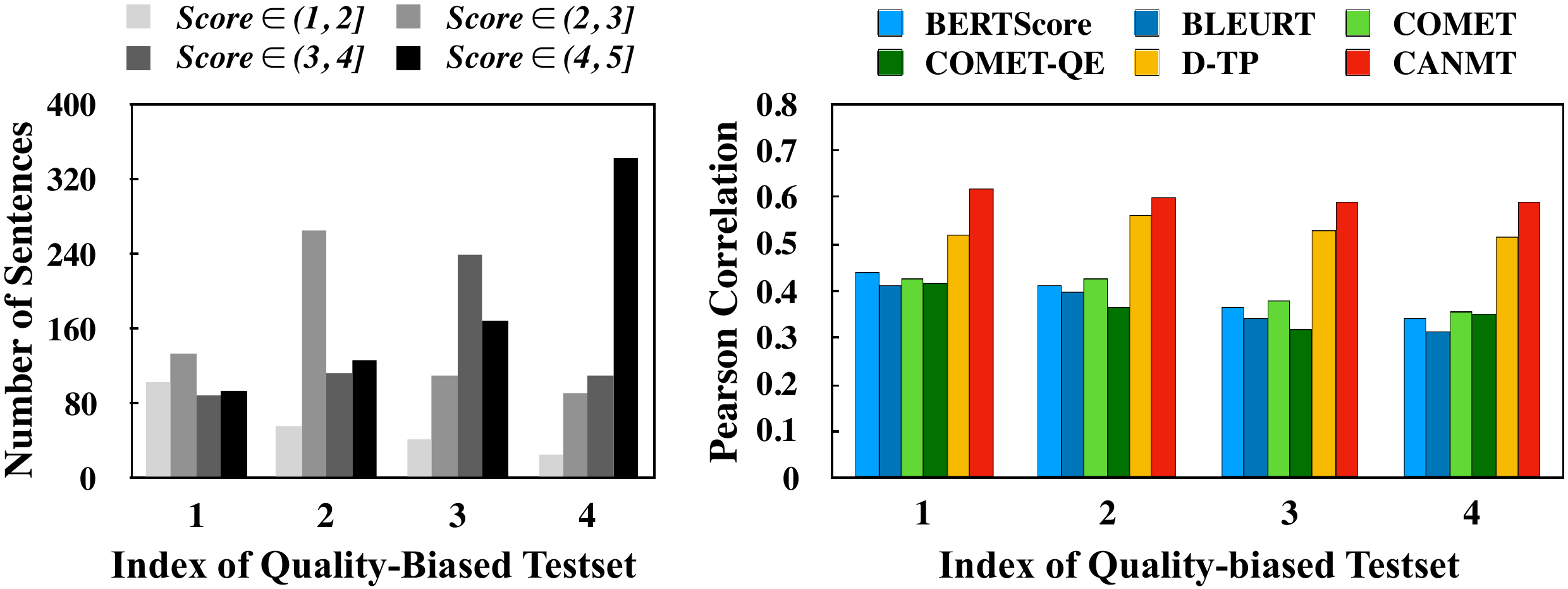}
         \caption{Results on Multi-Domain.}
         \label{fig:quality_drift_multidomain}
     \end{subfigure}
        \caption{Robustness on different quality-biased test sets. Four quality-biased tests (quality level improves from 1 to 4) are built according to the translation quality judged by linguists (left in each sub-figure). We show the Pearson correlation on different tests (right in each sub-figure). CANMT can better handle the problem of quality drift.}
        \label{fig:quality_drift}
\end{figure*}

\begin{figure}
\centering
\includegraphics[width=0.4\textwidth]{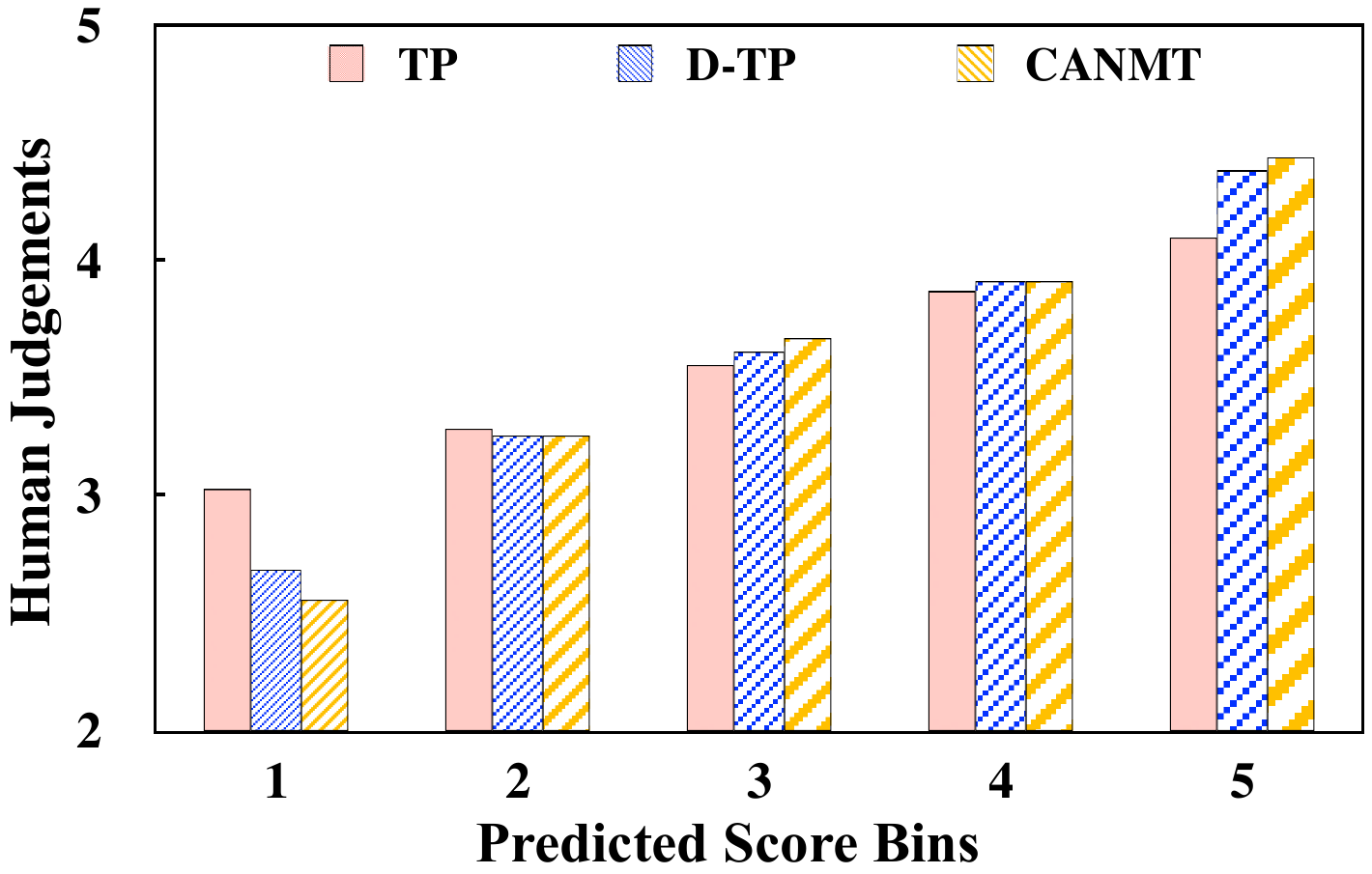}
\caption{Analysis on miscalibration. The x-axis represents five bins that divide translation samples according to predicted scores in ascending order. The y-axis indicates the average human score in each bin. 
}
\label{fig:confidence}
\end{figure}

\section{Analysis}
In this section, we conduct analyses on competency awareness to explain why CANMT performs better.

\begin{figure*}
     \begin{subfigure}[b]{0.312\textwidth}
         \includegraphics[width=\textwidth]{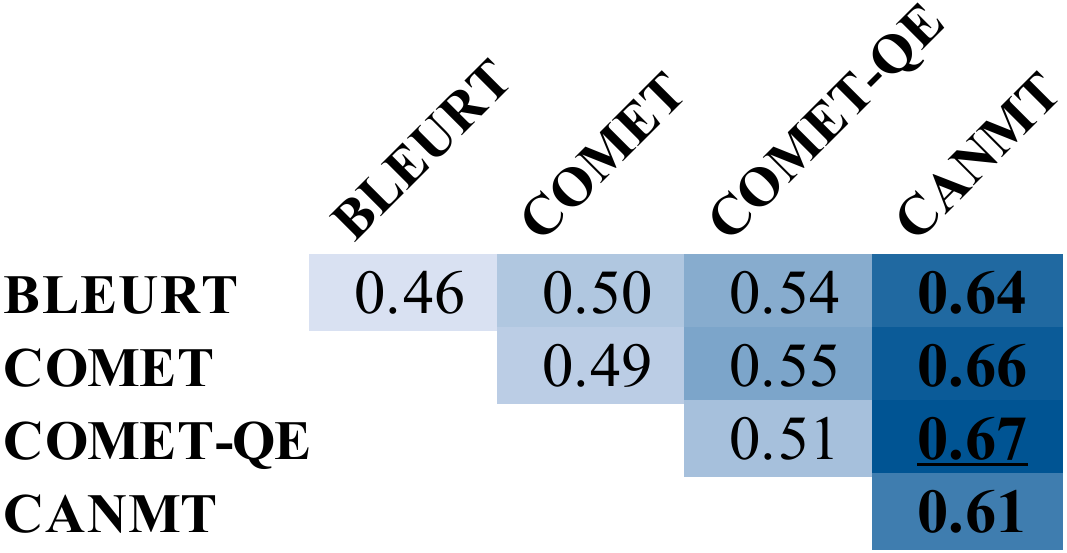}
         \caption{Zh$\to$En}
         \label{fig:comple_zh2en}
     \end{subfigure}
     \begin{subfigure}[b]{0.224\textwidth}
         \includegraphics[width=\textwidth]{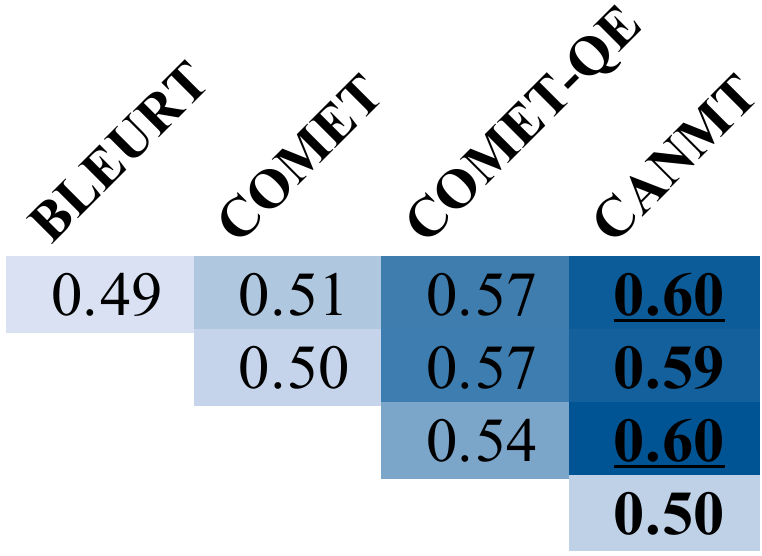}
         \caption{Fr$\to$En}
         \label{fig:comple_fr2en}
     \end{subfigure}
     \begin{subfigure}[b]{0.224\textwidth}
         \includegraphics[width=\textwidth]{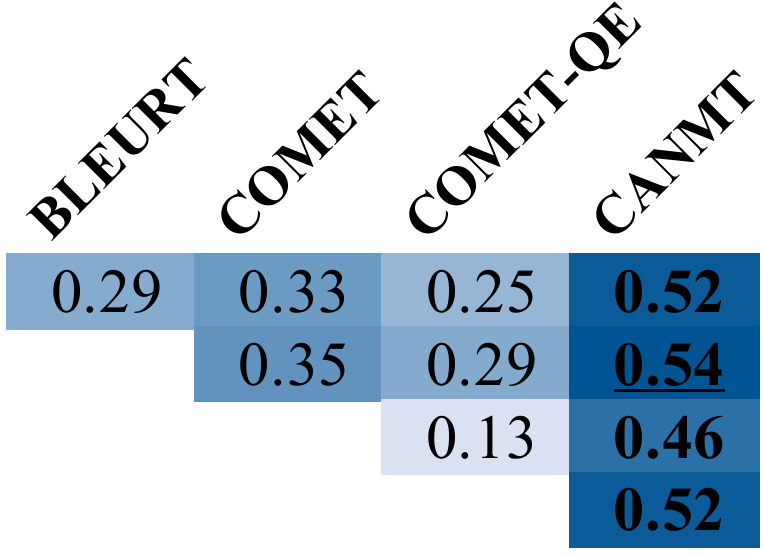}
         \caption{Ja$\to$En}
         \label{fig:comple_ja2en}
     \end{subfigure}
     \begin{subfigure}[b]{0.224\textwidth}
         \includegraphics[width=\textwidth]{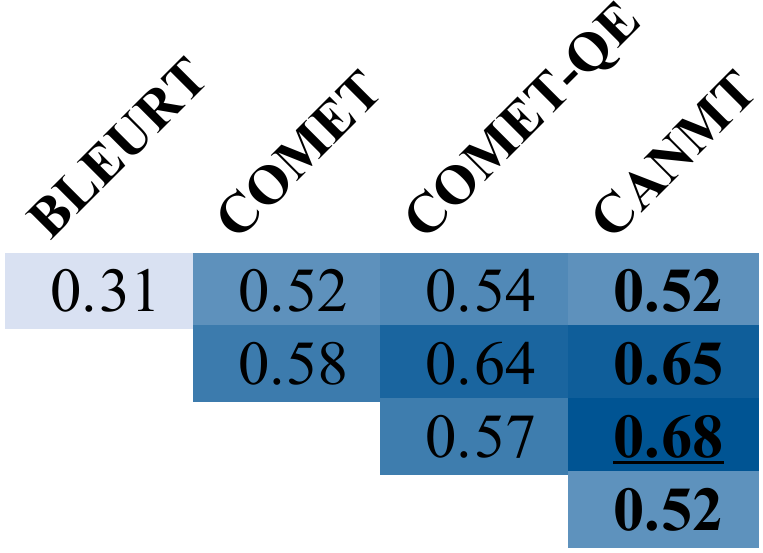}
         \caption{En$\to$De}
         \label{fig:comple_en2de}
     \end{subfigure}
        \caption{Complementary between different methods, where values in the cells of diagonal are the Pearson correlations of a single method, those of upper triangular matrix are the combination of two methods. Best values are underlined. As seen, ours and  extra-estimation methods are, to some extent, complementary to each other.}
        \label{fig:complementary}
\end{figure*}

\subsection{Robustness under Quality Drift} \label{robustness_quality_drift}
In order to verify whether CANMT behaves stably under different levels of translation quality, we draw on the quality-biased evaluation method in previous work of text generation evaluation\cite{Guan2020UNIONAU}.
We create four quality-biased test sets from Zh$\to$En task
by sampling human scores from four quality levels, each ranging $(1,2], (2,3], (3,4],(4,5]$ 
to represent
quality from poor to good.
The distribution (on the left) and correlation (on the right) for each quality-biased testset 
are illustrated in Figure~\ref{fig:quality_drift_news} (news test set) and Figure~\ref{fig:quality_drift_multidomain} (the hybrid multi-domain test set).
The Pearson scores of our method are stable across all quality-biased tests.
The main reason lies in the fact that the self-estimator of CANMT is trained jointly with NMT and also takes advantage of more decoding information from the MT model, which makes it aware of its own expertise and weaknesses.

\subsection{Analysis on Miscalibration}
\label{section:miscalibration}

It is known that neural networks suffer from the miscalibration problem between accuracy and confidence~\cite{guo2017calibration,wang2020inference}. 
Over-confidence can lead to predicting high confidence for poor translations, while under-confidence leads to low confidence for good translations.
Here, we evaluate whether \canmt alleviates the over- and under-confidence problem on Zh$\to$En multi-domain tests. For each method, we split sentences into five bins according to the predicted scores (ranging from low quality to high) and compute the average human scores. We plot these values of \canmt and confidence-based QE methods (TP and D-TP) in Figure~\ref{fig:confidence}.

As seen, comparing with \canmt, TP and D-TP achieve much higher average human scores in bin 1 (with the lowest predicted scores) but much lower ones in bin 5, suggesting that these methods face more serious over- and under-confidence problems than ours. Our approach performs better because of using more informative features from NMT models and an effective reconstruction strategy that considers the fidelity of the source sentence.


\subsection{Complementary Effects with Extra-Estimation}

Our method has the advantage of being more robust and better aware of its own competency, while extra-estimation is able to capture the supervised knowledge of the labeled data.
One interesting question is whether they can be complementary.
To verify it,  we simply sum up the z-normalized scores predicted by any two different methods for each sentence. We plot all the results in Figure~\ref{fig:complementary}.

Firstly, we can see that almost all the combinations of \canmt and extra-estimation tools (metrics or QE) outperform both its corresponding two single methods. The best combination of \canmt and extra-estimation can outperform the best single method with 0.06/0.06/0.02/0.11 improvements on the four tasks. This supports the complementary effects between extra-estimation and our method. 

Secondly, we can find that on all the tasks, the best-performing combination of each task always consists of one extra-estimation method and our method, which also outperforms the combination of the top two single methods. For example, in En$\to$De task, the combination of \canmt (0.52)  and COMET-QE (0.57) outperforms the combination of COMET (0.58) and COMET-QE (0.57) with 0.04, although the latter two methods are stronger single methods. This indicates the higher complementary effects between extra-estimation and \canmt are not simply explained by the ensemble of two strong methods. It is worth further research on how to integrate supervised information of human-scored data into competency awareness. 

\section{Related Work}
\label{section:related_work} 

Our work is related to studies in the translation evaluation community. 
As it currently stands, machine translation is generally evaluated by extra-estimation tools, including reference-based methods (metrics) and reference-less methods (QE). 

\subsection{Metric}
\label{section:metric} 
Existing metrics can be roughly categorized into two types: unsupervised metrics like BLEU~\cite{papineni2002bleu}, METEOR~\cite{lavie2009meteor} CHRF~\cite{popovic2015chrf}, and BERTScore~\cite{zhang2019bertscore}, and supervised metrics like ESIM~\cite{mathur2019putting},
BLEURT~\cite{sellam2020bleurt},  COMET~\cite{rei2020comet} and UniTE~\cite{Wan2022UniTEUT}. 
The evaluation ability of a metric relies on the given ground-truth references, which, however, are unavailable in a real-world application scenario.
Moreover, despite achieving strong performance, most of the state-of-the-art metrics are resource-heavy, relying on large-scale pre-trained language models and a significant amount of in-domain labeled data for training.
The findings of \newcite{Freitag2021ResultsOT} highlight that all state-of-the-art metrics exhibit performance drops when switching to out-of-domain test sets.
In contrast to these methods, CANMT exploits the internal information from the MT model to enable self-estimation of competency. Via decoupling the dependence on the human translation or expert-annotated data, our method is marginally affected by domain drift.

\subsection{Quality Estimation}
\label{section:qe} 
Different from Metrics, QE methods estimate translation quality without ground-truth  references.
There exist traditional feature-based approaches~\cite{Specia2009EstimatingTS} and neural network based methods~\cite{kepler2019openkiwi,fonseca2019findings,specia-etal-2020-findings-wmt} that exploit large scale language models and supervised on human-annotated data.
\newcite{Moon2020RevisitingRT} proposes an unsupervised method (RTT) that reconstructs source sentences by another backward MT and then estimates translation quality in terms of the lexical or semantic similarity between the original source sentence and its reconstruction through BLEU or BERTScore.
\citet{wang2019improving} and ~\citet{fomicheva2020unsupervised} propose confidence-based unsupervised QE methods that employ the model confidence to estimate quality with
Monte Carlo Dropout. 
Nevertheless, these methods merely consider the predicted distribution, regardless of the fidelity with respect to source sentences and the internal information of NMT. This makes confidence-based method exist certain miscalibration (over- and under-confidence) between translation accuracy and confidence~\cite{guo2017calibration,wang2020inference,Wan2020SelfPacedLF}. Besides, the Monte Carlo sampling requires repeatedly inference decoding and is thus heavily time-consuming.
\subsection{Reconstruction}
Reconstruction strategy has been widely used in machine translation~\cite{He2016DualLF,Liu2016AgreementOT,Cheng2016SemiSupervisedLF,Tu2017NeuralMT,Yee2019SimpleAE} to improve translation performance.
However, we employ reconstruction to equip machine translation with self-estimation.
Unlike previous studies that simply use an individual backward NMT for reconstruction, 
we make the self-estimator a part of the NMT model and exploit the continuous representations as input from the NMT decoder for reconstruction, which performs better in reflecting the quality of the translation.

\section{Conclusion}
In this paper, we pay attention to the competency awareness of NMT.
The major contributions of our work are three-fold: 1) We believe that competency awareness should be an essential capability of NMT and requires careful study. We hope our study can attract more interest in this problem; 2) We propose CANMT,  which is able to output both the competency and the translation for a given source sentence. It novelly leverages the internal information of itself to estimate the competency, showing superiority compared to other approaches that merely consider the outputs or the prediction distribution of NMT; and 3) Empirical results exhibit the strong performance and robustness of CANMT on quality estimation, which, to the best of our knowledge, is the first report that self-estimation method surpasses supervised ones across translation directions and domains.

For future work, it is a promising direction to employ competency awareness in reinforcement learning for NMT.
The self-estimated competency scores correlate better with human judgments than traditional reinforcement learning rewards such as sentence-level BLEU~\cite{Wu2018ASO}, we can use it to find weaknesses in NMT and take appropriate optimization.  


\section*{Limitations}

The limitations of CANMT are mainly twofold.
Firstly, the self-estimation performance of CANMT can only be evaluated by human translators at present, and there is a lack of an automated evaluation method, which may limit its further exploration.
Secondly, as we focus on the competency awareness of MT, we only verify the effectiveness of the model in estimating its own translation quality, and it is unknown for estimating the quality of translations generated by other NMT models.

\section*{Ethics Statement}
All procedures performed in studies involving human participants were in accordance with the ethical standards of the institutional and/or national research committee and with the 1964 Helsinki declaration and its later amendments or comparable ethical standards. This article does not contain any studies with animals performed by any of the authors. Informed consent was obtained from all individual participants included in the study.

\bibliography{anthology,custom}
\bibliographystyle{acl_natbib}

\appendix

\section{Dataset and Processing Strategies}
\label{appendix:dataset}

\begin{table}[h]
\centering
\small
\resizebox{0.5\textwidth}{!}{
\begin{tabular}{lrrrrrrrr}
\hline
\bf \multirow{2}{*}{Corpus} &
\bf \multirow{2}{*}{Language} &
\bf \multirow{2}{*}{Domain} &
\multicolumn{3}{c}{\bf Data Size}\\

& & &train & dev & test  \\
\hline
ASPEC & Ja$\to$En  & Science & 2M & 1790 & 1812  \\
CWMT17 & Zh$\to$En  & News & 9.2M & 2002 & 2001  \\

WMT14 & Fr$\to$En  & News & 35.7M & 26854  & 3003 \\

WMT14 & En$\to$De  & News & 4.5M & 3000 & 3003  \\
\hline
\end{tabular}}
\caption{Number of parallel sentences in each dataset. ``M'' stands for ``million''.}
\label{tab:data}
\end{table}
The corpus size and processing details are shown in Table~\ref{tab:data}.
The byte-pair encoding~\cite{Sennrich2016NeuralMT} for Zh$\to$En, Fr$\to$En, En$\to$De, and Ja$\to$En are trained respectively with 32k/40k/32k/16k merge operations and  source-target vocabularies are shared except for Zh$\to$En.

\section{Baseline Details}
We use the official code for each quality estimation tool. For BERTScore, the officially recommended F1 score is adopted.
BERTScore-en based on the pre-trained model RoBERTa-large is used for to-English translation tasks, while BERTScore-multi based on the pre-trained model BERT-base is used for En$\to$De.
The evaluated metrics BLEURT and COMET are pre-trained on BERT-base and XLM-RoBERTa-large respectively, and then finetuned on WMT15-18 Metric rating datasets(for BLEURT) and WMT17-19 Metric Direct Assessment datasets(for COMET).

\begin{table*}[t]
\begin{tabular}{c|l|l}
\toprule
\bf Score & \bf Accuracy & \bf Fluency \\
\midrule
5/Excellent &  & Accurate grammar and words. Fluent and idiomatic. \\
\cline{1-1} \cline{3-3}
\multirow{3}{*}{4/Good} &  & The grammar and words are accurate.\\
 & The translation (basically) & The translation is relatively fluent with only a \\
 & faithfully reflects the meaning & few errors that do not affect understanding. \\
\cline{1-1} \cline{3-3}
\multirow{4}{*}{3/Acceptable} & of the original text. &  The grammar is basically accurate, but there  \\ 
& & are some  problems, such as inaccurate words, \\ 
& & improper collocation, and insufficient fluency, \\
& & which can be fixed at a small cost. \\
\hline
\multirow{2}{*}{2/Poor} &   & The syntax is basically accurate, but there \\
& The translation deviates from & are obvious semantic losses or errors.\\
\cline{1-1} \cline{3-3}
\multirow{2}{*}{1/Very bad} & the original meaning. & Difficult to read or understand. The translation \\
& & is irrelevant to the semantics of the original text. \\
\bottomrule 
\end{tabular}
\caption{Standard of Human Evaluation.}
\label{tab:human_eval}
\end{table*}

\section{Standard of Human Evaluation}
\label{appendix:human_eval}
The human evaluation of translation quality mainly focuses on two aspects: accuracy and fluency.
Details about human scoring for translation quality are elaborated in Table~\ref{tab:human_eval}. 
Each sentence is independently annotated by three qualified translators from professional language service providers. We train the annotators before annotations. Any evaluation in which the variance of scores is greater than 0.5 will be rechecked to ensure consistency. We use the average score given by three experts as the final human judgment.
The human annotation results are released in the supplemental material.

\section{Computational Cost}
According to inference, the translation cost is the same as Transformer, while the self-estimation is equal to the forced decoding of the Transformer Decoder. For example, the QE inference costs (in seconds) for the 200-sentence Fr→En test set of each method are as follows: CANMT 5 $<$ BLEURT 15 $<$ BERTScore 31 $<$ COMET-QE 42 $<$ COMET 43 $<$ D-TP 70. The computation of training can be represented as much as twice of a standard Transformer.
\end{document}